\documentclass{article}

\usepackage[final]{nips_2018}

\usepackage[utf8]{inputenc} % allow utf-8 input
\usepackage[T1]{fontenc}    % use 8-bit T1 fonts
\usepackage{hyperref}       % hyperlinks
\usepackage{url}            % simple URL typesetting
\usepackage{booktabs}       % professional-quality tables
\usepackage{amsfonts}       % blackboard math symbols
\usepackage{nicefrac}       % compact symbols for 1/2, etc.
\usepackage{microtype}      % microtypography
\usepackage{enumitem}

\usepackage{graphicx}
\usepackage{amssymb}
\usepackage{bbold}

\usepackage[]{algorithm2e}

\usepackage{amsmath}
\usepackage[table, x11names]{xcolor}
\colorlet{lightgrayX50}{lightgray!50}

\newcommand{\degree}{^\circ}

\usepackage[para,online]{threeparttable}

\usepackage{makecell}

\usepackage{multirow}
\usepackage{tabto}

\makeatletter
\renewcommand{\paragraph}{%
  \@startsection{paragraph}{4}%
  {\z@}{0.2ex \@plus 0.2ex \@minus .2ex}{-1em}%
  {\normalfont\normalsize\bfseries}%
}
\makeatother

\newcommand{\Image}{\mathcal{I}}
\newcommand{\argmax}{\mathop{\mathrm{argmax}}}

\title{Correction of Electron Back-scattered Diffraction datasets using an evolutionary algorithm}

\author{
  Florian Strub$^*$\\
  Univ. Lille, CNRS, Centrale Lille,\\ Inria, UMR 9189 - CRIStAL\\
  \texttt{florian.strub@inria.fr} \\
  \And
  Marie-Agathe Charpagne\thanks{Equal contribution}\\
  Univ. of California, Santa Barbara,\\ Materials Department \\
  \texttt{marie.charpagne@engineering.ucsb.edu} \\
  \And
  Tresa M. Pollock \\
  Univ. of California, Santa Barbara,\\ Materials Department \\
  \texttt{pollock@engineering.ucsb.edu} \\
}

\begin{document}

\maketitle

\begin{abstract}
In materials science and particularly electron microscopy, Electron Back-scatter Diffraction (EBSD) is a common and powerful mapping technique for collecting local crystallographic data at the sub-micron scale. The quality of the reconstruction of the maps is critical to study the spatial distribution of phases and crystallographic orientation relationships between phases, a key interest in materials science. However, EBSD data is known to suffer from distortions that arise from several instrument and detector artifacts.
In this paper, we present an unsupervised method that corrects those distortions, and enables or enhances phase differentiation in EBSD data. The method uses a segmented electron image of the phases of interest (laths, precipitates, voids, inclusions) gathered using detectors that generate less distorted data, of the same area than the EBSD map, and then searches for the best transformation to correct the distortions of the initial EBSD data. To do so, the Covariance Matrix Adaptation Evolution Strategy (CMA-ES) is implemented to distort the EBSD until it matches the reference electron image. Fast and versatile, this method does not require any human annotation and can be applied to large datasets and wide areas, where the distortions are important. Besides, this method requires very little assumption concerning the shape of the distortion function. Some application examples in multiphase materials with feature sizes down to 1 $\mu$m are presented, including a Titanium alloy and a Nickel-base superalloy.

\end{abstract}

\section{Introduction}
\label{sec:introduction}

\begin{figure}[t]
\centering
\includegraphics[width=0.4\textwidth]{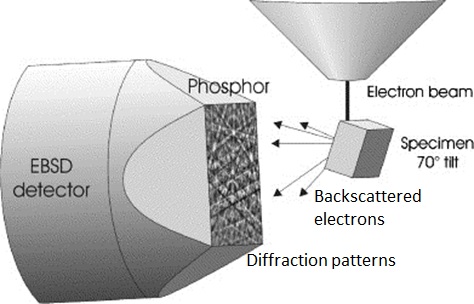}
\caption{Schematic representation of an EBSD setup, from Wilkelmann et al. \cite{Winkelmann2007}}
\label{fig:EBSD}
\vskip -1em
\end{figure}

Electron Backscattered Diffraction (EBSD) is a powerful tool for gathering local crystallographic data in multiphase crystalline materials. Automated, it enables the acquisition of large datasets. This technique enables a fast mapping of microstructures with a good angular resolution (typically 0.5$\degree$). Widely used in scanning electron microscopy (SEM), it requires careful preparation of the surface to be mapped, by mirror polishing and slight chemical etching for enhanced diffraction. A typical setup is shown in Figure \ref{fig:EBSD}. In the SEM, an electron beam rasters the surface of interest, which is tilted at an angle of 70$\degree$. The electron beam interacts with the material, producing back-scattered electrons (among others), which can be diffracted. 
%Resulting from various interactions of the incident electron beam with the material, several types of electrons are produced. Some electrons enter the matter and are then back-scattered. They may escape from the surface with an angle that corresponds to a Bragg condition, related to the spacing of the planes of the crystal they interacted with. 
The resulting diffraction pattern is composed of bands ("Kikuchi" bands) which are collected using a CCD camera. These bands correspond of the planes of the crystal and a careful analysis of the diffraction bands in those patterns enable to trace down the nature and orientation of the crystals at each (X,Y) point in the map. Therefore, it is critical to obtain the precise location of those orientations.
Unfortunately, EBSD data suffers from many distortion phenomena that arise from instrument and detector artifacts \cite{Nolze2007}. The most common distortions are the following: \textit{First order distortions:} the maximum diffraction contrast is reached when the surface to analyze is tilted to an angle of 70$\degree$. In practice, the samples are mounted on stages which are then tilted. If the rotation axis of the stage is not perfectly parallel to the surface of the sample, an area that would be a square at 0$\degree$ tilt turns into a trapezoid after tilting. Those distortions can be approximated by an affine transformation, which components are a translation and a rotation. \textit{Second order distortions:} in the microscope, the electron beam is deflected by a set of lenses that induce a barrel distortion on the final image, mostly visible at low magnification, which can be modeled by a second-order polynomial function. \textit{Drift distortions:} when a sample is being exposed for some time to the electron beam, charges accumulate on its surface, leading to the deflection of the beam. This effect occurs in any scanning process and is usually the most pronounced at the beginning of the scan \cite{Zhang2014}. The amount of beam drifting varies from a sample to another and depends on all the scanning parameters. Given the complexity of the net resulting distortion, it cannot be easily calculated and undistorting EBSD data requires strong (and limiting) physical hypothesis.

%F: We need to give a few (unsuccessful method) and describe them in 

%Another limitation of the method can be the differentiation of phases. Because of the similarity of the diffraction patterns they produce, some couples of phases are intrinsically challenging to distinguish with this method. This is the case for $\gamma$ and $\gamma '$ phases (both of cubic symmetry) in Nickel and Cobalt-base superalloys \cite{Charpagne2016}.
%F Sauf erreur de ma part, c'est un autre probleme en fait que tu ennonce ici? Hough transformations ne permettent pas de differencier certainses phases à cause du contraste. Cela est différent de corriger les distortions, non?

Payton and Nolze \citep{Payton2013} have proposed a method that involves the acquisition of Back-Scattered Electron (BSE) images on a tilted sample, using specific detectors. While they demonstrate the capability of that method for segmenting small features precisely, the limit of this technique is that not all the EBSD detectors are equipped with such built-in diode sensors.
Another way to tackle both problems of distortions %and phase differentiation
is to gather corresponding data over the same area, using detectors that are subjected to reduced distortion phenomena, and exhibiting some phase contrast. This is the case of Secondary Electrons (SE) and/or Backscattered Electrons (BSE) images, collected at a 0$\degree$ tilt angle (i.e., which have no distortion). Some methods have been implemented using this principle \cite{Zhang2014}. However, they require the manual selection of a set of carefully selected matching points in the image and the EBSD data. The higher the number of pairs of points, the greater the precision. This step can be time-consuming or challenging due to the potential lack of similar contrasted features in both EBSD and BSE data.

In this paper, we propose to use speckles of similar features without any human annotations, allowing the fast and precise recombination of the data. To do so, we combine a high number of automatically raw generated reference points in both the EBSD and the undistorted BSE data by using physical properties of the speckles such as microstructural features such as pores, a second-phase, precipitates as shown in Fig~\ref{fig:R65_all} and Fig~\ref{fig:Ti64all}. We then define a superimposition score between the speckles and look for an undistort function that maximizes this metric. As a second contribution, we propose a method that relies on generating artificially distorted meshes that are used to regress an undistort function. Those meshes are iteratively generated by using a black-box optimizer, namely, Covariance Matrix Adaptation Evolutionary Strategy (CMA-ES)~\cite{Hansen1996,Hansen2016}, to slowly refine the undistort function. We can then correctly locate the nature and orientation of the crystals at each (X,Y) point in the map.
We finally assess our method on two difficult EBSD case examples: a Titanium alloy and a Nickel-base superalloy while studying the stability and reproducibility of our approach.
 
%%%%%%%%%%%%%%%%%%%%%%%%%%%%%%%%%%%%%%%%%%%%%%%%%%%%

\section{Method}
\label{sec:method}

The method we present here enables compensation of the distortions in EBSD data and adds phase differentiation if necessary, by accurately matching it to BSE images collected at individual pixel precision and over broad areas. To do so, the Covariance Matrix Adaptation Evolution Strategy (CMA-ES)~\cite{Hansen1996,Hansen2016} is used to calculate the relative distortion function between the EBSD data and the BSE image. The code source, the hyperparameters and one data set are available at \url{https://github.com/MLmicroscopy}.

\subsection{Defining a similarity measure between the speckles}
\label{par:measure}

In order to superimpose the EBSD and BSE speckles, a quantitative measurement of their similarity is needed. The Dice similarity metric \cite{Dice1945} (also known as F1-score)  describes the relative overlap between the segmented images. More precisely, this score measures the overlap between segmented pixels. Formally, given the binary EBSD speckle $\Image^{EBSD} \in \{0,1\}^{I \times J}$ and the binary BSE image $\Image^{BSE} \in \{0,1\}^{I
\times J}$ where $1$ encodes the segmented object and $0$ the background image, and ${I \times J}$ their domain of definition, the Dice similarity for is computed as follows:
\begin{align}
	\text{Similarity}(\Image^{EBSD}, \Image^{BSE}) = 2\frac{|\Image_{seg}^{EBSD} \cap \Image_{seg}^{BSE}|}{|\Image_{seg}^{EBSD}| + |\Image_{seg}^{BSE}|}
    \label{eq:similarity}
\end{align}
where $|.|$ encodes the size of the underlying set and $\Image_{seg}^{.}$ only corresponds to the segmented pixels in the image. Thus, $|\Image^{EBSD}_{seg} \cap \Image^{BSE}_{seg}|$ corresponds to the number of matching pixels and $|\Image^{EBSD}_{seg}|$ and $|\Image^{BSE}_{seg}|$ respectively correspond to the total number of pixels in the EBSD and BSE images. This similarity measure counts the number of overlapping segmented pixels normalized by the sum of segmented pixels. It ranges from 0 (no overlapping segmented pixels) to 1 (perfect matching).  The value of the ${Similarity function}(\Image^{EBSD}, \Image^{BSE})$ is referred to as the 'score' in the following. As the Dice similarity is intuitive, robust and easy to compute, it has been extensively used in several image segmentation applications such as medicine~\cite{Taha2015}.

\subsection{Correction work-flow}
Starting from the initial EBSD data and the BSE image, the two speckles are generated. In a first step (initial alignment), the BSE speckle is rescaled so its pixel size corresponds to that of the EBSD data. The rescaling is done using a nearest neighbor interpolation. The BSE speckle image is also rotated and translated so it is pre-aligned to the EBSD data. To do so, a grid-search over a set of affine transformations is performed, to distort the re-sized speckle over three parameters: two translations along the x and y directions, and a rotation. The user is free to choose the widths of the translation (within steps down to one pixel) and the rotation angle (with a precision as fine as wanted). All the possible translations and rotations within the provided ranges are tested. The transformation leading to the best score is saved and applied to the BSE speckle. The latter is then used in the second step, as the reference speckle for the compensation of the distortions. In the second step, the EBSD speckle is meshed on a regular grid, and the mesh is distorted using the CMA-ES optimizer, in an iterative process. The initial and new locations of the points of the mesh are used as pairs of points, to estimate the distortion function. At each iteration, the score produced by the distorted EBSD speckle is calculated. The distortion function corresponding to the distorted speckle that exhibits the best score is used to fill in the crystallographic information in the new EBSD file, and the segmented BSE image is used to fill in the phase data (if needed). The parametrization and calculation of the distortion function are explained in the following.

\subsection{The distortion function}
\label{subsec:distor}

Formally, the goal is to find the distortion function $f$ that maximizes the similarity between the EBSD and the BSE speckles. This leads to the optimization procedure described in eq. \ref{eq:optimisation}:
\begin{align}
	f^* &= \argmax_f \Big( \text{Similarity}(\Image^{EBSD}, \Image^{BSE}) \Big) \\   
	&= \argmax_f  \Big( 2\frac{|f(\Image^{EBSD})_{seg} \cap \Image_{seg}^{BSE}|}{|f(\Image^{EBSD})_{seg}| + |\Image_{seg}^{BSE}| } \Big)
    \label{eq:optimisation}
\end{align}
As described in section~\ref{sec:introduction}, many overlapping distortion phenomena occur while collecting the EBSD data, leading to a non-linear distortion. In the following, the nature of the distortion function is assumed to be polynomial. It is a very weak assumption as polynomial function can approximate a large range of complex functions. Besides, the proposed method can be easily extended to other function approximators such as gaussian processes, trees etc. Formally, a polynomial distortion of degree $P$ is defined as follow:
\begin{align}
 f(x,y) =
  \begin{cases}
	x' = \sum_{n=0}^{P} \sum_{k}^{n} c^x_{k,n-k}x^{k}y^{n-k}   \\
	y' = \sum_{n=0}^{P} \sum_{k=0}^{n} c^y_{k,n-k}x^{k}y^{n-k}
\end{cases}
  \label{eq:polynomial_fit}
\end{align}
%Higher order polynomial fits can be computed by following a binomial distribution over the (x,y) exponents. 
Given a polynomial order, the goal is to find the optimal set of weights $c^*$ that maximizes the similarity measure. This set of weights $c$ can be computed by two means: by directly looking into the space of parameters $c$ or by generating intermediate matching points that are used to perform a polynomial regression. 

While the first approach may be more direct, the manifold of parameters $c$ is very difficult to explore, as a small change of parameters can lead to drastically different distortions. Therefore, in the proposed method, a set of matching points is generated, that are used to regress the polynomial coefficients $c$. As a small change in the location of the matching points leads to a close distortion, this process allows a better granularity in exploring the space of the distortions. More precisely, a regular mesh and a distorted one are generated, and then the coefficients $c$ are determined by a least-squares fitting method that maps the distorted mesh to the regular one. This function is described below:
\begin{itemize}
    \item Generate a distorted mesh $M = ( (x_0, y_0), \ldots, (x_N, y_N) )$
    \item Generate a regular mesh $M' = ( (x_0', y_0'), \ldots, (x_N', y_N,) )$
    \item Find the distortion $f$ parametrized by $c$ such as $c = \text{argmin}_{\hat{c}} \sum_i^N ||\bigl( \begin{smallmatrix} x_i'\\y_i' \end{smallmatrix} \bigr) - f_{\hat{c}}(x_i, y_i) ||_2 $
\end{itemize}

where, $||.||$ is the euclidean norm. Note that the dimension of the mesh is tuned by the user via the number of points $N$ and/or the step-size between points. The influence of that parameter is discussed in paragraph \ref{sec:discussion}. As a result, finding the optimal distortion $f^*$ requires to find the optimal set of matching points (or mesh) $M^*$ that describes the underlying distortion the best. These matching points are then used to estimate the distortion itself. Counter-intuitively, the distorted mesh may not match physical reality as discussed further in the discussion~\ref{subsec:repeatability}

Finally, the \textit{polynomial distortion} function is called from the python package \textit{Sklearn} \cite{scikit-image} and \textit{Skimage} \cite{scikit-image}. The next paragraph describes how the pairs of matching points are created, using the CMA-ES optimizer.

\subsection{Using CMA-ES as a randomized black-box optimization to correct the distortions}
The goal of the optimization is to find the function $f$ that maximizes the $Similarity$ function between the speckles by finding the best distorted mesh $M$ (which is non-differentiable). To do so, a black-box optimizer method is used, CMA-ES, which stands for Covariance Matrix Adaptation Evolutionary Strategy. Black-box optimization algorithms are efficient at solving non-linear, non-convex optimization problems, in continuous domains. They also overcome the deficiencies of the derivative-based methods in complex multidimensional landscapes that are rugged, noisy, have outliers or local optima, or when no error-gradient is available. All those features make the black-box optimizer relevant for the present problem, where the shape of the distortion is unknown and the process of superposing speckles intrinsically implies the $Similarity $ function to reach local maxima. 

CMA-ES has become a standard tool for continuous optimization and has been applied in various fields of research, such as image recognition for biology \cite{Ibanez2009,Sisniega2017}, energy \cite{Reddy2013}, chemistry \cite{Fateen2012,Weber2015}. Though, to the authors knowledge, it has not been applied in the field of electron microscopy.
%CMA-ES is also suitable for search spaces that include up to 100 dimensions \cite{Hansen2016a} which is the case

CMA-ES belongs to the family of evolutionary strategies (ESs), they are iterative algorithms, based on the principles of natural selection. CMA-ES involves a parametrized distribution (a multivariate normal distribution) that evolves throughout the iterations~\cite{Hansen1996,Hansen2016}. 
ESs follow several steps: initialization, sampling, evaluation and update. 
The initialization step consists in initializing the probability distribution parameters and to sample a initial population of individuals accordingly. At each iteration, a new \textit{generation} of individuals is created. Those individuals are candidate solutions to the optimization problem, whose goodness can be evaluated on a \textit{fitness function}. After evaluation of the fitness of each individual in that population, the statistics of the distribution are updated according to the algorithm at hand. A new population is re-sampled according the new distribution and the process is repeated until a termination criterion is reached. Those operations are described below:

\begin{algorithm}[H]
 Initialize the distribution parameters $\theta$ \\
 \For{generation $g=0,1,...number\ of \ iterations$}{
 Sample $\lambda$ individuals $x_{1}, x_{2}, ..., x_{\lambda} \in \mathbb{R}^{n}$ according the probability distribution $P_{\theta}(x)$\\
    Evaluate them on the fitness function \\
    Update $\theta \leftarrow F(\theta, x_{1}, x_{2}, ..., x_{\lambda}, Similarity(x_{1}), Similarity(x_{2}), ..., Similarity(x_{\lambda}))$ \\
 }
\end{algorithm}

where $P_{\theta}$ is a probability distribution that describes where the good solutions are believed to be and $F(.)$ is some update rule. In the CMA-ES, $P_{\theta}$ is a multivariate normal distribution \cite{Hansen1996,Hansen2016}. Those steps are detailed in the following.

\paragraph{Initialization} The distribution $P_{\theta}$  encodes the distribution of the distorted spatial mesh. More precisely, the distribution is a multi-variate normal distribution where each dimension encodes either the x-coordinate or y-coordinate of a point of the mesh. For example, if the mesh is a 4x4 grid, the multi-variate normal distribution has 32 dimensions (16 points with 2 coordinates). The initial distribution is defined by centering the mesh distribution on a regular grid. The standard deviation then encodes the initial acceptable moving distance of the matching points. Both the dimension of the mesh and the standard deviation are chosen by the user. In practise, CMA-ES is going to slowly and iteratively distort the regular mesh, in order to retrieve a set of matching points that encode the distortion.

\paragraph{Sampling} New distorted meshes are generated following the distribution $P_{\theta}$. The coordinates are rounded, and matching points outside the range of the speckle dimension are kept. 

\paragraph{Evaluation} The fitness function is the $Similarity$ described in eq. \ref{eq:similarity}. the coefficients $c$ of the function are first regressed by matching the distorted mesh to the regular one. The similarity measure is then computed and returned as the fitness score.

\paragraph{Update}
The basic CMA equation for sampling the search points at the next generation is given in eq. \ref{eq:gaussian_mutation}.
\begin{align}
x_{k}^{(g+1)} \leftarrow m^{(g)} + \sigma^{(g)} \mathcal{N}(0, \mathcal{C}^{(g)}),\ for\ k=1, 2, ..., \lambda.
  \label{eq:gaussian_mutation}
\end{align}
where $x_{k}^{(g+1)} \in \mathbb{R}^{n}$ is the $k-th$ offspring of the generation $g+1$. $m^{g} \in \mathbb{R}^{n}$ is the mean value of the distribution at the generation $g$. $\sigma^{(g)} \in \mathbb{R}^{+}$ is the step size at the generation $g$. $\mathcal{N}(0, \mathcal{C}^{(g)})$ is a multivariate normal distribution with a mean of zero and covariance matrix $\mathcal{C}^{g}$. $\mathcal{C}^{g} \in \mathbb{R}^{n x n}$ is the covariance matrix of the distribution at the generation $g$. It is symmetric definite positive and describes the geometrical shape of the distribution. The initial value of $\sigma$, $\sigma_{0}$, is picked by the user and $\mathcal{C}_{0} = \mathcal{I}$. 
They both evolve throughout the iterations, as the population evolves. The self-adaptation of those parameters is the key point for the rapid convergence of the optimization~\cite{Hansen2016}. The equations governing the update of $\sigma^{(g)}$, $\mathcal{C}^{(g)}$ and $m^{g}$ are described in~\ref{App:CMA_update}. 
In the present case, this update shifts the spatial distribution of the points of distorted mesh in order to search for the optimal distortion function. 

\paragraph{Termination} This process is repeated until the termination criterion is reached. In the present case, a criterion based on the maximum number of steps is used, but more advanced methods can be used to detect when the optimization starts plateauing. If several speckles are known to have the same distortion, they can be used as a validation criterion to avoid over-fitting. Once the CMA-ES algorithm is finished, the means of the distribution $P_{\theta}$ are used as the final distorted mesh.

\subsection{Generation of the new EBSD map}
Once the maximum number of calls has been reached, the polynomial function leading to the best superimposition of the speckles is applied to the EBSD data. In this process, the original grid of the EBSD is kept and the Euler angles, confidence indexes and other useful data, are interpolated using the polynomial function. The points of the grid that end up containing no data are filled with zero-values for the confidence indexes and the $(0,0,0)$ triplet of Euler angles. The rotated and translated segmented BSE image is used to fill in the phase data of the new EBSD file. For that reason, and as already mentioned, the segmentation of the BSE image has to be made as accurately as possible. An application example is shown in section \ref{sec:results}.

%% Results
\section{Results}
\label{sec:results}
Samples of two materials have been characterized in a FEI Versa 3D SEM, equipped with a TSL EDAX EBSD Hikari Plus Camera with an indexing speed of 300 to 400 frames per second with 4x4 binning. The computer for the post-processing was equipped with a Quad Core processor, 4.2 GHz and 64 GB of RAM. Most computations required about 10 to 15 minutes.

\subsection{$\gamma$ and $\gamma '$ phases in Rene 65 superalloy}
Rene 65 is a polycrystalline Nickel-based alloy that has been designed for turbine disk applications. The surface of the sample was prepared using conventional polishing techniques, followed by vibratory polishing using a 0.04 $\mu m$ $Al_{2}O_{3}$ suspension. Its microstructure consists of fine $\gamma$ matrix grains, which average equivalent diameter is 10-12 $\mu m$ and spherical primary $\gamma '$ precipitates which are located on the grain boundaries; their equivalent diameter is in the range of 1-4 $\mu m$. Those phases exhibit different crystallographic structures which however lead to similar diffraction patterns, making them undistinguishable on EBSD data. The goal of the reconstruction here was to both differentiate the phases and to correct the distortions. EBSD maps and corresponding BSE images have been acquired using an acceleration voltage of 20 kV with a step size of 0.1 $\mu m$ over an area of 150 x 200 $\mu m$. The CMA-ES optimizer was used on a mesh grid of 25 x 25 points, an initial standard deviation of 20 pixels and a polynomial order of 3.
\begin{figure}[t]
\centering
\includegraphics[width=1\textwidth]{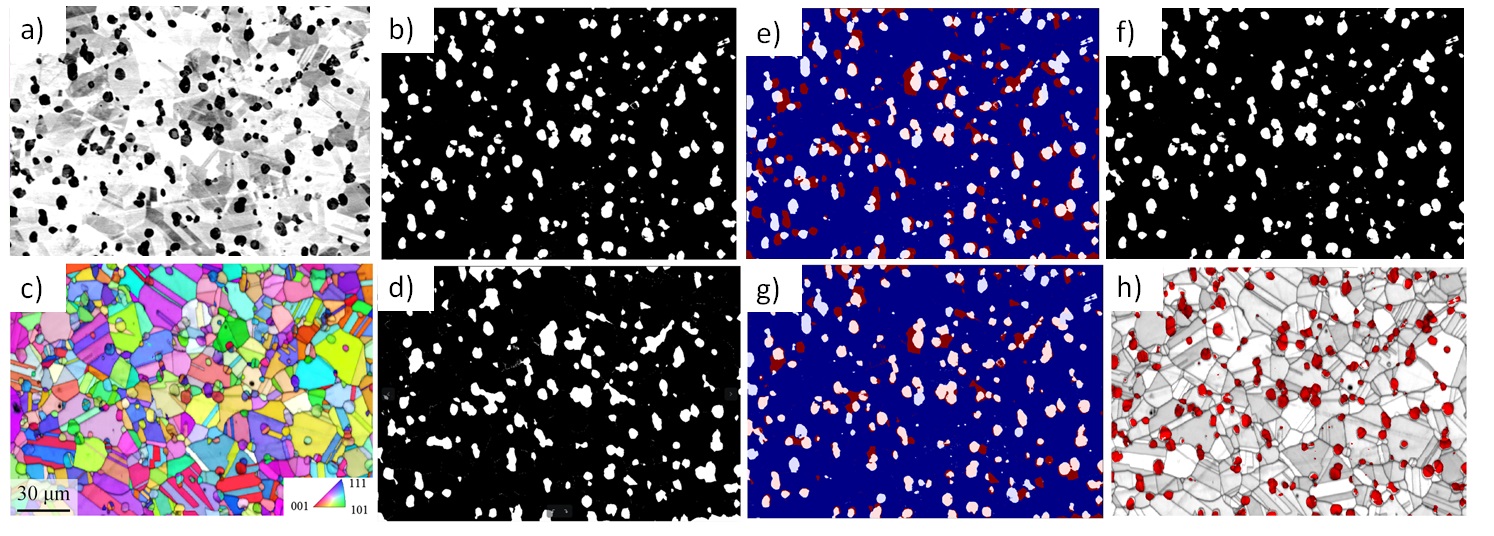}
\caption{Nickel-base superalloy dataset: a) BSE image, b) segmented BSE image, c) initial EBSD dataset (colored according to the orientation of the crystals projected along the normal to the polished surface), d) EBSD speckle, e) speckle superimposition after alignment, f) aligned segmented BSE image (reference), g) speckle superimposition after applying the CMA-ES optimizer, h) final phase map with the precipitates colored in red, showing a good spatial reconstruction of the two phases.}
\label{fig:R65_all}
\vskip -1em
\end{figure}

Figure \ref{fig:R65_all} shows the process of data correction. The input data consists of a BSE image (fig. \ref{fig:R65_all}-a) that is segmented to create the BSE speckle (fig. \ref{fig:R65_all}-b) on one hand, and EBSD data (fig. \ref{fig:R65_all}-c) from which a second speckle is generated (fig. \ref{fig:R65_all}-d). In this particular case, the precipitates have been chosen for the speckles. The EBSD speckle has been generated by segmenting the smallest features of the map, which is a very inaccurate way to segment the precipitates, yet sufficient for the compensation of the distortions as will be shown later. Fig. \ref{fig:R65_all}-e shows the superimposition of the speckles after rescaling and alignment of the BSE speckle. Some precipitates -those located in the center of the area mostly- match, while those located away from the center do not superimpose. Fig. \ref{fig:R65_all}-f shows the aligned BSE speckle which serves as a reference for the CMA-ES optimization. The CMA-ES optimizer plateaued at a final score of 0.62, leading to a good superimposition of the speckles, as shown on fig. \ref{fig:R65_all}-g. The final EBSD map is displayed on fig. \ref{fig:R65_all}-h, where the grayscale map reveals the confidence of the indexation of the diffraction patterns (thus, crystal boundaries appear in dark grey), and the precipitates are colored in red. This map illustrates the good matching between the location of the crystal boundaries and the location of the precipitates. 

\subsection{$\beta$ phase in $\alpha - \beta$ Titanium alloy}
 A sample of Ti-6Al-4V with an $\alpha - \beta$ structure was embedded in bakelite and electro-chemically polished. It was then mounted on the SEM stage and copper tape was used to enable electrical conductivity from the sample holder to the polished surface. An area of 14 x 20 $\mu m$ has been characterized with a 20 kV acceleration voltage and a step size of 40 nm. One purpose of collecting this dataset is to better resolve the phase map, using the better quality of the segmentation of the BSE image, with a high precision and fine details. This mounting system was purposely used in order to induce a large drift of the beam during the EBSD scan.
 %\begin{figure}[t]
 %\centering
 %\includegraphics[width=0.7\textwidth]{figures/Ti64comparison.jpg}
 %\caption{Speckle superposition after correction of a) Affine distortions only, b) All distortions. On a blue background,% the EBSD speckle is colored in red and the BSE speckle in white.}
% \label{fig:Ti64comparison}
% \end{figure}
% \subsection{$\beta$ phase in $\alpha - \beta$ Titanium alloy}
% \label{Ti64}

 \begin{figure}[t]
 \centering
 \includegraphics[width=0.95\textwidth]{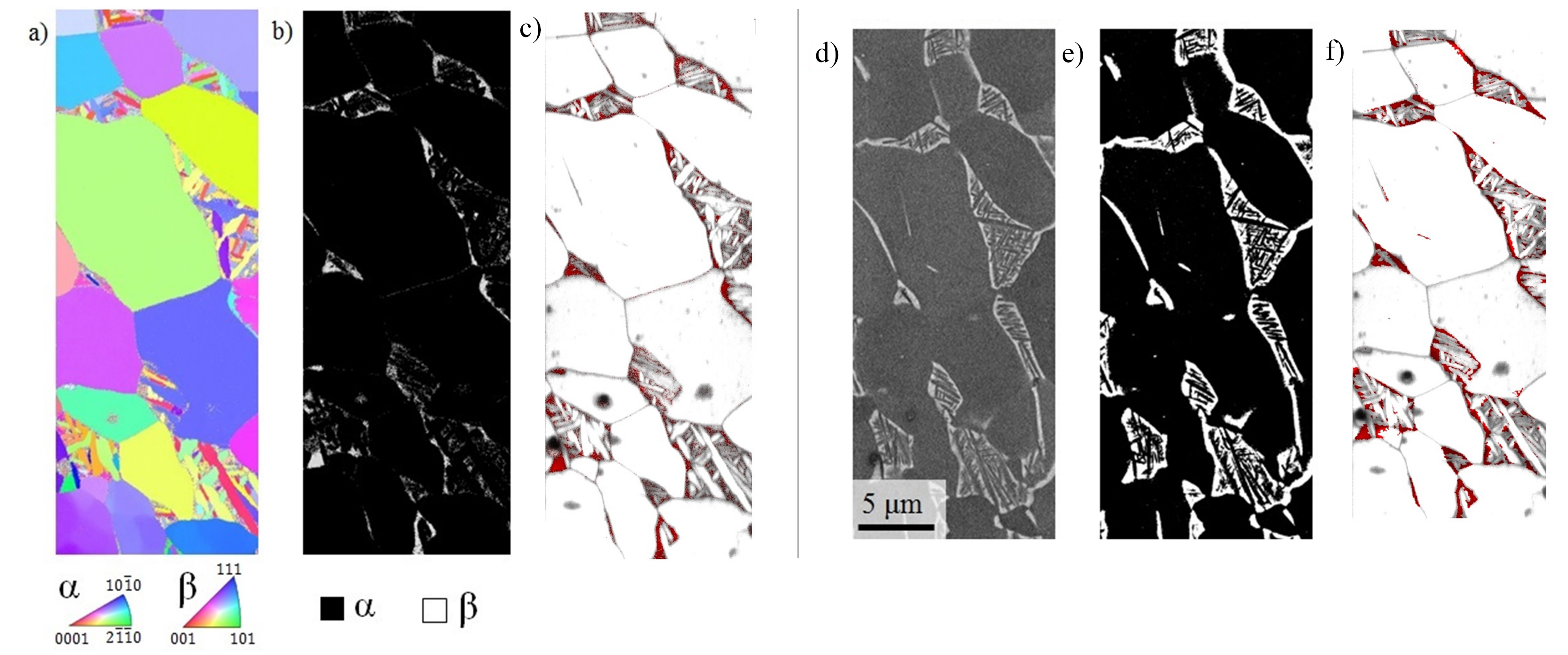}
 \caption{Microstructure of the Ti-64 sample: a) Orientation map colored following the Inverse Pole Figure color code (projected along the normal to the sample surface), b) Phase map with $\alpha$ phase in black and $\beta$ phase in white (EBSD speckle), c) Corresponding BSE image, d) Thresholded BSE image. Phase reconstruction in Ti-6Al-4V, the phase map with transparent $\alpha$ phase and red-colored $\beta$ phase is superposed to the Image Quality map: e) Initial phase map, f) Final phase map.}
 \label{fig:Ti64all}
 \vskip -1em
 \end{figure}

 The orientation map for this sample is shown on fig. \ref{fig:Ti64all}-a. The corresponding BSE image is shown on fig. \ref{fig:Ti64all}-d, where the $\beta$ phase exhibits a lighter contrast than the $\alpha$ phase. Fig. \ref{fig:Ti64all}-b shows the initial phase map with the $\beta$ phase in white. This phase map is superimposed to the Index Quality map on fig. \ref{fig:Ti64all}-c, where the $\beta$ phase is colored in red. The comparison of fig. \ref{fig:Ti64all}-b and -e shows that only a small fraction of the $\beta$ phase has been identified as such on the initial EBSD data. Its initial area fraction is 3.4\%, versus 9.9\% in the segmented BSE image. The CMA-ES optimizer has been applied on a 25 x 25 points mesh grid, with an initial standard deviation $\sigma_{0}=20$ pixels, using a polynomial order of 3 and 5,000 iterations. Fig. \ref{fig:Ti64all}-f shows the final phase map, colored as fig. \ref{fig:Ti64all}-c. It superimposes well with the Index Quality map. The final area fraction of $\beta$ phase is 9.9\%, as in the BSE speckle. The contribution of the CMA-ES optimizer to compensate higher order distortions and match all the features of the speckles is discussed in section \ref{subsec:versatility}.

%% Discussion %%%
\section{Discussion}
\label{sec:discussion}

\subsection{Versatility of the method}
\label{subsec:versatility}

\begin{figure}[t]
\centering
\includegraphics[width=0.90\textwidth]{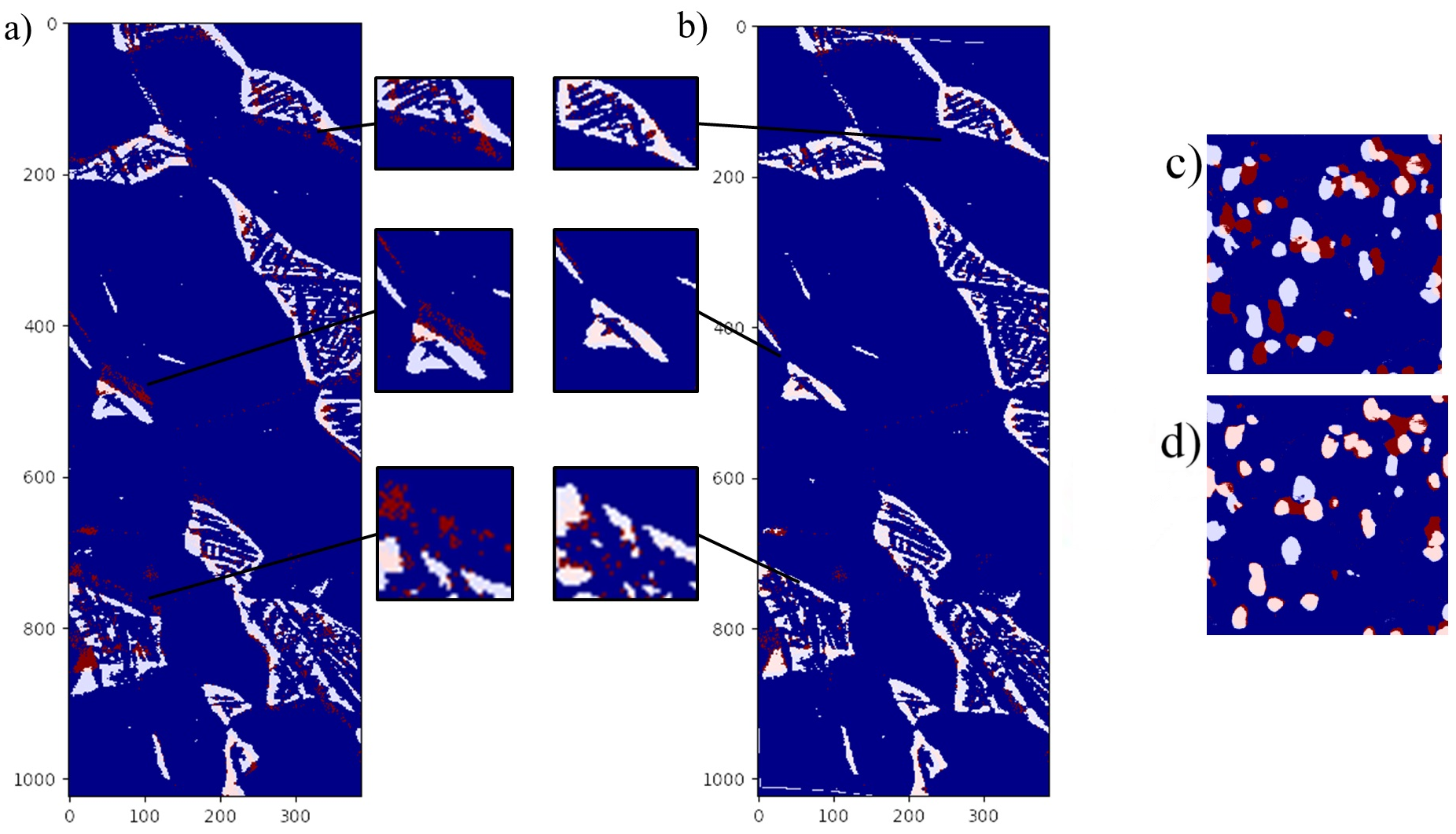}
\caption{Speckle superposition for the Titanium alloy a) after correction of affine distortions only, b) all distortions, and for the Nickel alloy c) affine distortions, d) all distortions. On a blue background, the EBSD speckle is colored in red and the BSE speckle in white.}
\label{fig:R65comparison}
\vskip -1em
\end{figure}

Several methods have been proposed in the literature, for correcting the distortions in EBSD data, using various algorithms to make up for the drift distortions. Zhang et al. \cite{Zhang2014} have shown that a thin plate spline function enables compensation of the distortions in EBSD data. Contrary to most methods, the use of the CMA-ES strategy does not assume any specific shape of the distortion function, nor constrain the order of the polynomial function. The use of speckles and a score to quantify the goodness of the superposition enables the matching of both images at the resolution of the pixel. Fig. \ref{fig:R65comparison} shows a comparison between the superposition of the speckles after the compensation of affine distortions only (a, c) and after the CMA-ES procedure (b, d). On a blue background, the EBSD speckle is colored in red and the BSE speckle in white. While many features did not even superimpose after the initial alignment, the CMA-ES optimizer was able to compensate the finer distortions -even very local and over the whole area- and achieve a better matching of the speckles despite an important drift of the beam.

\subsection{Convergence, stability and repeatability of the CMA-ES optimization}
\label{subsec:repeatability}
During the CMA-ES optimization procedure, new distorted meshes are generated to regress the polynomial distortion. Counter-intuitively, these meshes are not constrained to encode a realistic distortion, they are only optimized such that the final distortion is meaningful. Thus, one must not use the distorted meshes outside the polynomial regression step, as several different meshes can lead to the same distortion function. Indeed, there exists an infinite number of sets of points that extrapolate to the same final function. Those points do not have to always be on the curve itself. To a lesser extent, several polynomial coefficients may also lead to a very similar resulting distortion function, only differing around the border of the speckle.

Despite this apparent limitation, CMA-ES turns out to be highly reproducible in practice. As the score increases, the step-size $\sigma$ decreases with the number of iterations, indicating little variation around the mesh ground.

\begin{figure}[t]
\centering
\includegraphics[width=0.95\textwidth]{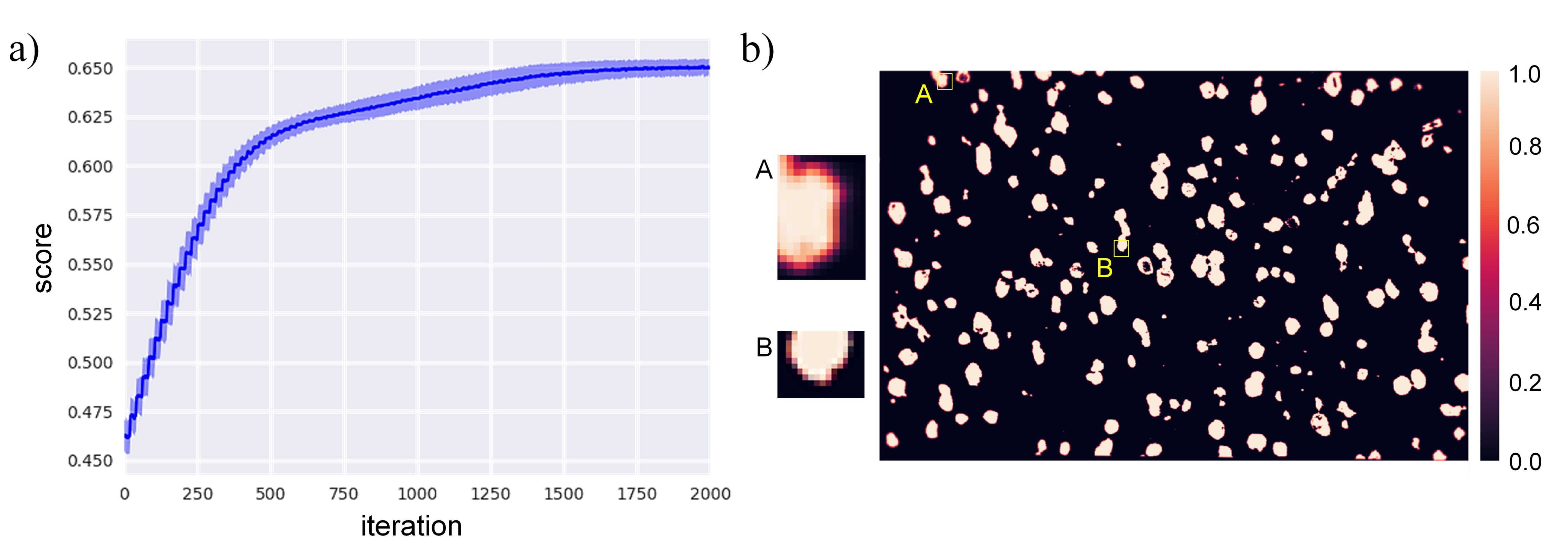}
\caption{Convergence and stability of the CMA optimization over 100 runs. a) Mean score (dark blue), minimum and maximum values (light blue), as a function of the number of iterations,  b) Corresponding heatmap : the pixels are colored according to the number fraction of times they were identical among the 100 runs.}
\label{fig:repetition}
\vskip -1em
\end{figure}

The CMA-ES strategy has been applied 100 times on the same dataset. Each run consisted in 2000 iterations, with a step size of 75 points, an initial standard-deviation $\sigma_{0}$ of 5 pixels. A polynomial order of 3 was used for the polynomial function. The CMA optimizer produced an average score of 0.6587 with an associated standard deviation of 0.0015. The similarity score between distorted speckles over the 100 runs was 0.9353 in average, indicating a good precision and repeatability of the process. Figure \ref{fig:repetition}-a shows the evolution of the score throughout the iterations: the dark blue curve corresponds to the mean score over 100 runs. The light blue colored area around the curve corresponds to the lower and upper bounds of the score (mean - standard deviation, and mean + standard deviation, respectively). Those values show that the optimization is stable and repeatable over several runs. Some steps are clearly visible on the first 500 iterations and correspond to the new bounds determined by the creation of new individuals at each generation in CMA.

Based on those statistics, a heat map has also been generated, displayed on fig. \ref{fig:repetition}-b. This map shows the segmented BSE speckle colored according to the number fraction of times that a pixel was present at a given (x', y') location. In other words, a pixel that appears in white (value 1.0) was always assigned to the $\gamma '$ phase. On the opposite, a pixel that appears in black was never assigned this phase. The more consistent and precise the optimization, the sharper the contrast on this map. The shape of the precipitates appears clearly on fig. \ref{fig:repetition}-b, which is consistent with the good repeatability suggested by fig. \ref{fig:repetition}-a. This map also gives information about the precision of the reconstruction over the whole map. The center of the precipitates usually appears with a value of 1.0, however their boundaries are usually less precise. The deviation on the reconstruction of the boundaries of the precipitates is illustrated on two examples, labeled A and B. The precipitate A is close to the border of the image, and the precipitate B is in the center. The inserts on the side of the figure show that the boundaries of those two precipitates are not reconstructed with the same consistency over the 100 slices. There is only one layer of pixels having a value lower than 1.0 on the precipitate B, versus 2 to 3 rows on the boundary of the precipitate A. The boundaries of most of the precipitates located on the edges of the map have a similar coloring. This indicates that the CMA optimizer leads to consistent results in the center of the map, with a deviation of about 1 pixel as the location of the phase boundaries are approached. At the borders of the map, the consistency on the location of the phase map is only about 2 to 3 pixels. On this specific dataset, the associated error on the location of the phase boundaries is 0.2 to 0.3 $\mu m$, which remains much smaller than the actual size of the features of interest. In other words, the reconstruction does not artificially create additional features (precipitates) nor assign the wrong phase to any feature.

\section{Conclusion and perspectives}
\label{sec:conclusion}

A new method for the compensation of the distortions and improved phase differentiation in EBSD data has been developed. The principle consists in using an electron image of the same area than the EBSD data, taken at a 0$\degree$ tilt angle, and using it as a reference. Similar features are segmented out of the EBSD data and electron image. Then, the CMA Evolutionary Strategy is applied in order to match the speckles. The goodness of the superposition is measured by a score, which ranges from 0 to 1. The extrapolated distortion function is then applied to the EBSD data and crystallographic and phase data are recombined in a new EBSD file.

This method has been applied successfully to a nickel-based superalloy containing two phases, where it has enabled to separate the phases while compensating the distortions. Supposing a precise segmentation of the phases on the electron image, this method can reach a precision of a couple of pixels over broad areas, despite important drift phenomena.
This method can be used on any EBSD dataset, as long as two speckles of similar features can be generated.

\section*{Acknowledgements}
The authors gratefully acknowledge a Vannevar Bush Fellowship, ONR Grant N00014-18-1-3031. \newline
We acknowledge the following agencies for research funding and computing support: CHISTERA IGLU and CPER Nord-Pas de Calais/FEDER DATA Advanced data science and technologies 2015-2020
Henry Proudhon from Mines ParisTech is acknowledged for prolific conversations and exchange of ideas. 
Olivier Pietquin, J\'er\'emie Mary and Philippe Preux from Inria are acknowledged for insightful conversations. 
Xuedong Shang from INRIA is acknowledged for his various comments on CMA-ES. 
McLean Echlin from the University of Santa Barbara is acknowledged for providing the Ti-6Al-4V sample. 
Mickael Kirka from the Oak Ridge National Laboratory is acknowledged for providing the additive manufactured Inconel 718 sample. 
The Carlton Forge Works company (PCC Corporation) is acknowledged for providing the Rene 65 material.

%\newpage

{\small
\bibliographystyle{plain}
\bibliography{main}
}
\appendix

\section{Update of the parameters in CMA-ES}
\label{App:CMA_update}
Set $m \in \mathbb{R}^{n}$, $\sigma \in \mathbb{R}+, \lambda$ \hfill Input\\
$C = I, p_{c} = 0, p_{\sigma}=0,$ \hfill Initialize parameters\\
$c_{c} \approx 4/n, c_{\sigma} \approx 4/n, c_{1} \approx 2/n^{2}, c_{\mu} \approx \mu_{w}/n^{2}, c_{1} + c_{\mu} \leq  1, d_{\sigma} \approx 1+ \sqrt{\frac{\mu_{w}}{n}}, w_{i=1 ... \lambda}$ such that $\mu_{w} = \frac{1}{\sum_{i=1}^{\mu} w_{i}^{2}} \approx 0.3 \lambda$ \newline
While not terminate, \newline
$x_{i} = m + \sigma y_{i}$, where $y_{i} = \mathcal{N}_{i}(0, C),$ for $i = 1, ..., \lambda$ \hfill Sampling and variation\\
$m \leftarrow \sum_{i=1}^{\mu} w_{i}x_{i:\lambda} = m + \sigma y_{w}$, where $y_{w} = \sum_{i=1}^{\mu} w_{i}x_{i:\lambda}$ \hfill Update mean \\
$p_{c} \leftarrow (1 - c_{c}) p_{c} + \mathbb{1}_{\|p_{\sigma}\| < 1.5\sqrt{n}}\sqrt{1 - (1 - c_{c}^{2})}\sqrt{\mu_{w}}y_{w}$ \hfill Cumulation for the calculation of $C$ \\
$p_{\sigma} \leftarrow (1 - c_{\sigma})p_{\sigma}) + \sqrt{1-(1-c_{\sigma})^{2}} \sqrt{\mu_{w}} C^{-\frac{1}{2}} y_{w}$ \hfill Cumulation for the calculation of $\sigma$ \\
$C \leftarrow (1 - c_{1} - c_{\mu})C + c_{1}p_{c}p_{c}^{T} + c_{\mu} \sum_{i=1}^{\mu} w_{i}y_{i:\lambda}y^{T}_{i:\lambda}$, \hfill Update of the covariance matrix $C$\\
$\sigma \leftarrow \sigma exp(\frac{c_{\sigma}}{d\sigma} (\frac{\|p_{\sigma}\|}{E\| \mathcal{N}(0, I)\|} - 1))$ \hfill Update of the step size $\sigma$.\\

\end{document}